%% file: root.tex
\documentclass[letterpaper, 10 pt, conference]{ieeeconf}  

\IEEEoverridecommandlockouts                              

\usepackage{cite}
\usepackage{amsmath,amsfonts}
\usepackage{algorithmic}
\usepackage{array}
\usepackage[caption=false,font=normalsize,labelfont=sf,textfont=sf]{subfig}
\usepackage{textcomp}
\usepackage{stfloats}
\usepackage{url}
\usepackage{verbatim}
\usepackage{graphicx}
\usepackage{hyperref} 
\usepackage{balance}
\usepackage{booktabs}
\usepackage{multirow}
\usepackage{xcolor}
\usepackage{cleveref}
\usepackage{bm}
\usepackage{amsmath}
\usepackage{siunitx}
\usepackage{algorithm}
\usepackage{tcolorbox}
\usepackage[switch, columnwise]{lineno}

\title{\LARGE \bf
LaMMA-P: Generalizable Multi-Agent Long-Horizon Task Allocation and Planning with LM-Driven PDDL Planner
}

\author{Xiaopan Zhang$^{*}$, Hao Qin$^{*}$, Fuquan Wang, Yue Dong, and Jiachen Li$^{\ddag}$
\thanks{$^*$Equal contribution \ $^\ddag$Corresponding author}
\thanks{X. Zhang, F. Wang, Y. Dong, and J. Li are with the University of California, Riverside, USA. {\tt\small \{xzhan006, fuquanw, yued, jiachen.li\}@ucr.edu}}
\thanks{H. Qin is with Penn State University, USA. This work was done when H. Qin was a visiting student at the University of California, Riverside. {\tt\small hxq5039@psu.edu}
}
}

\begin{document}

\maketitle

\begin{abstract}
    \input{contents/00abstract}
\end{abstract}

\section{Introduction}\label{sec:introduction}
    \input{contents/01introduction}

\section{Related Work}\label{sec:relatedwork}
    \input{contents/02relatedwork}

\begin{figure*}[htbp]
	\centering
    \includegraphics[width=0.9\linewidth]{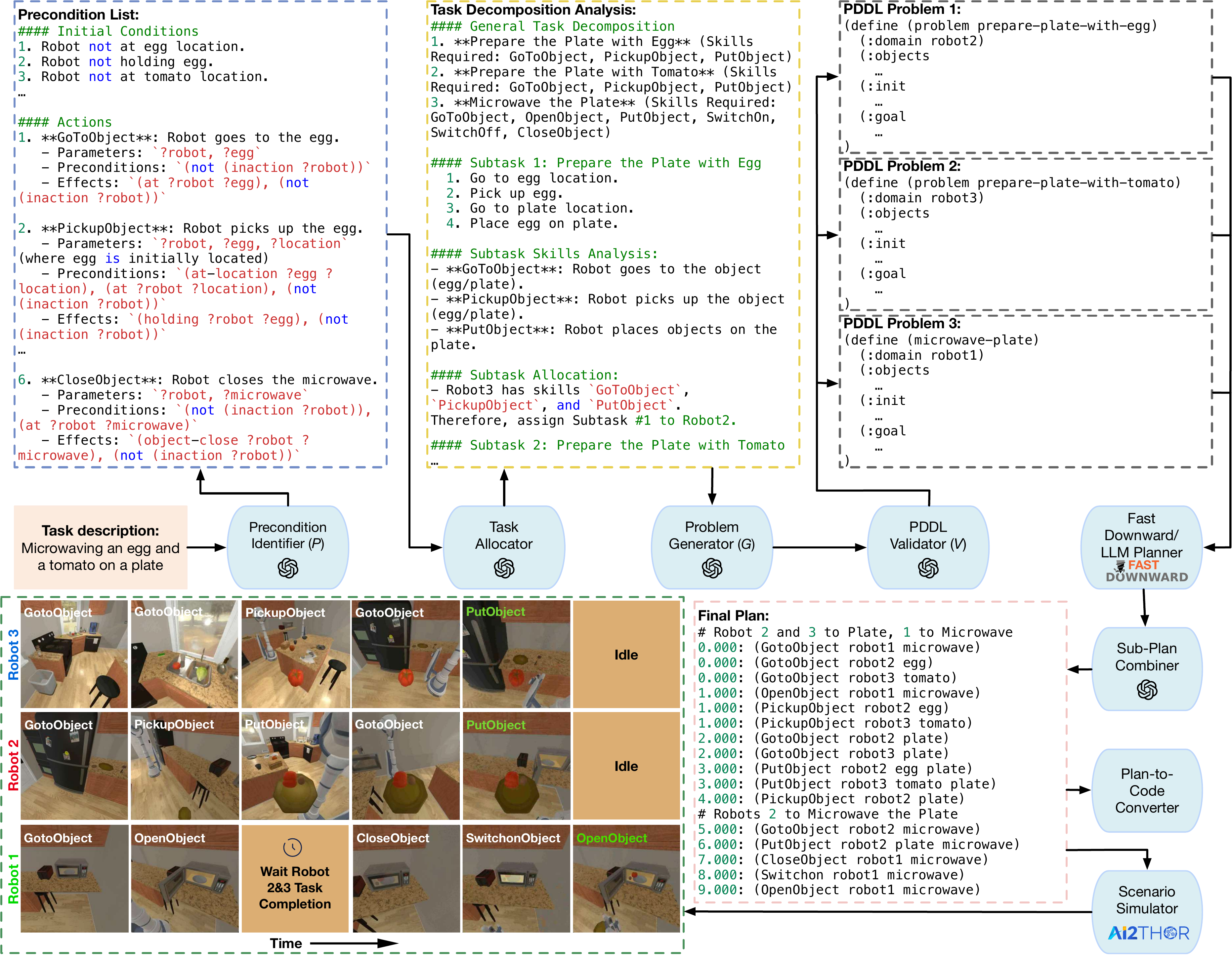}
    \vspace{-0.2cm}
	\caption{An overview of LaMMA-P’s modular architecture. Our framework leverages LMs within six key modules: \textit{Precondition Identifier} (\textit{P}), \textit{Task Allocator}, \textit{Problem Generator} (\textit{G}), \textit{Fast Downward/LLM Planner}, \textit{PDDL Validator} (\textit{V}), and \textit{Sub-Plan Combiner}, each serving a specific role in task execution. The \textit{Precondition Identifier} analyzes the initial conditions and requirements for task completion. The \textit{Task Allocator} assigns subtasks to agents based on their skill sets and task complexity. The \textit{Fast Downward/LLM Planner} converts task descriptions into executable plans for each agent. Finally, the \textit{Sub-Plan Combiner} integrates individual agent plans into a cohesive execution strategy, ensuring synchronized actions to achieve the overall task objectives.}
    \vspace{-0.5cm}
	\label{fig:framework}
\end{figure*}

\section{Problem Formulation}\label{sec:problemformulations}
    \input{contents/03problemformulations}

\section{Method}
\label{sec:methodology}
    \input{contents/04methods}

\section{Experiments}\label{sec:experiments}
    \input{contents/05experiments}

\section{Conclusion}\label{sec:conclusions}
    \input{contents/06conclusions}

\bibliographystyle{IEEEtran}
\bibliography{References}

\end{document}

%% file: contents/00abstract.tex
Language models (LMs) possess a strong capability to comprehend natural language, making them effective in translating human instructions into detailed plans for simple robot tasks. Nevertheless, it remains a significant challenge to handle long-horizon tasks, especially in subtask identification and allocation for cooperative heterogeneous robot teams. 
To address this issue, we propose a Language Model-Driven Multi-Agent PDDL Planner (LaMMA-P), a novel multi-agent task planning framework that achieves state-of-the-art performance on long-horizon tasks. 
LaMMA-P integrates the strengths of the LMs' reasoning capability and the traditional heuristic search planner to achieve a high success rate and efficiency while demonstrating strong generalization across tasks. 
Additionally, we create MAT-THOR, a comprehensive benchmark that features household tasks with two different levels of complexity based on the AI2-THOR environment. 
The experimental results demonstrate that LaMMA-P achieves a 105\% higher success rate and 36\% higher efficiency than existing LM-based multi-agent planners. The experimental videos, code, datasets, and detailed prompts used in each module can be found on the project website: \href{https://lamma-p.github.io}{https://lamma-p.github.io}.

%% file: contents/01introduction.tex
\begin{figure}[!tbp]
        \centering
        \includegraphics[width=0.95\linewidth]{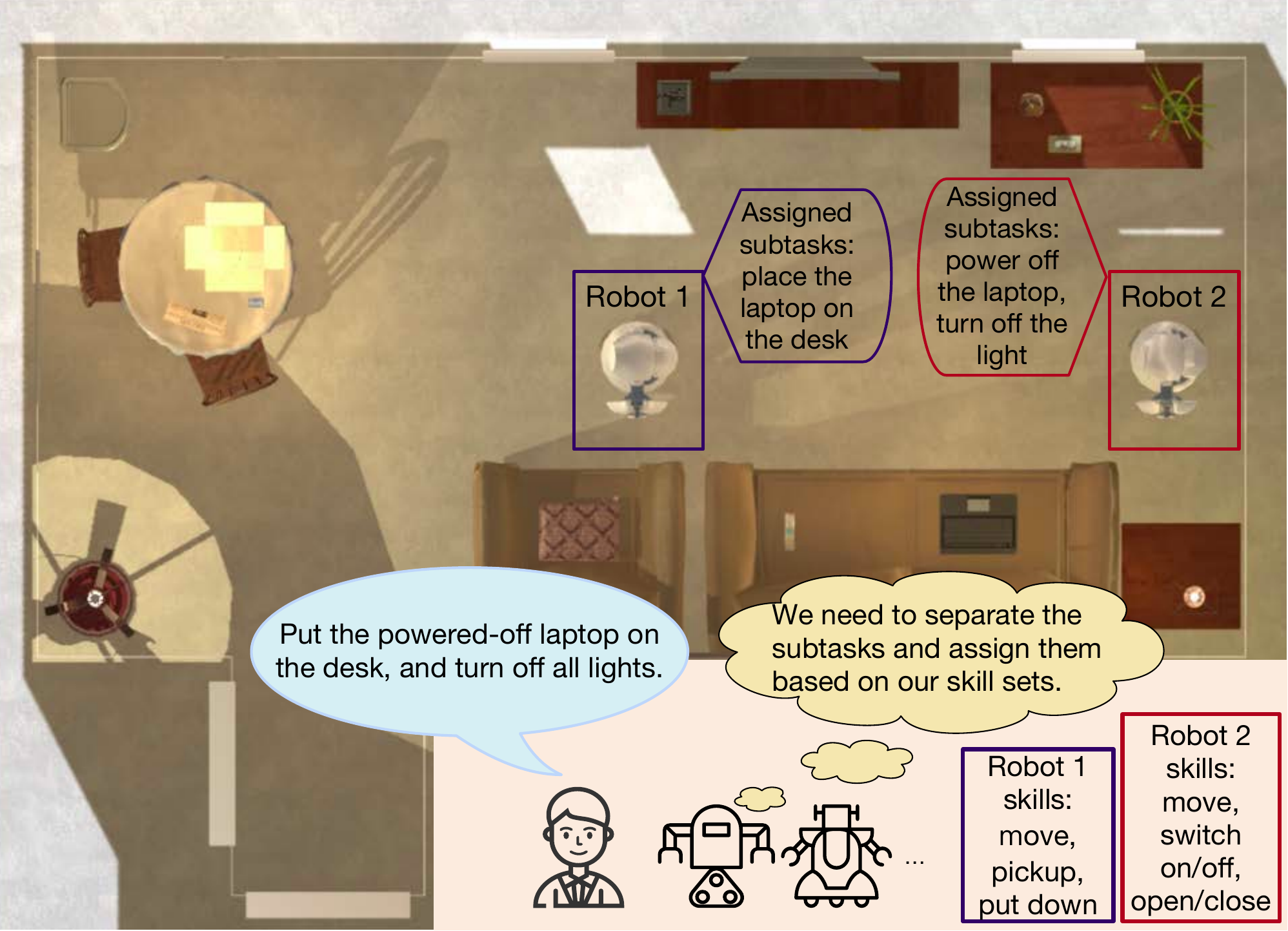}
        \vspace{-0.2cm}
	\caption{A typical multi-agent long-horizon task in a household scenario where Robot 1 and Robot 2 collaborate to execute tasks based on human commands given in natural language.}
        \vspace{-0.6cm}
	\label{fig:motivation}
\end{figure}

Multi-robot systems have been widely applied to various real-world tasks, such as search and rescue \cite{baxter2007multi, queralta2020collaborative}, warehouse automation, and agricultural processes \cite{bolu2021adaptive, oliveira2021advances, chen2024accounting}. These systems enable multiple robots to collaborate autonomously, which often have well-defined objectives and require efficient coordination between robots. 
Recently, language models (LMs) have been applied to complex, long-horizon household tasks, allowing robots to understand semantics and execute natural language commands \cite{zhang2023building, nayak2024long}. Fig.~\ref{fig:motivation} illustrates multi-robot cooperation in a household setting that exemplifies the complexity of long-horizon task planning, where robots need to perform a sequence of interconnected tasks that may require coordinated actions. 
Thus, different robots need to be assigned specific tasks based on their capabilities. 
For instance, Robot 1 is responsible for moving a laptop, while Robot 2 is tasked with switching off the lights. Such scenarios highlight the challenges of long-horizon planning for heterogeneous robot teams, including task generalization, efficient sub-task allocation, and ensuring optimal coordination. 
Managing these tasks over a long horizon requires both effectively leveraging each robot's skills and accurately identifying parallelizable tasks to maximize performance.

Traditional multi-robot task planning approaches have difficulty in managing such complex long-horizon tasks, especially in environments with diverse tasks and complex interdependencies between robots \cite{rizk2019cooperative, wang2022distributed}. 
They often rely on fixed, pre-defined algorithms that are not flexible enough to handle the intricacies of tasks that extend over a long duration \cite{khamis2015multi}. 
Although recent approaches \cite{kannan2023smart, wang2024safe, singh2024twostep, mandi2024roco} that leverage language models (LMs) for multi-agent task planning have shown potential, they often struggle with long input sequences and complex task dependencies, especially in collaborative, multi-robot settings. They also fall short in generalization across tasks with varying levels of difficulty.

To address these challenges, we propose {LaMMA-P}, a \textbf{La}nguage \textbf{M}odel-Driven \textbf{M}ulti-\textbf{A}gent \textbf{P}DDL Planner, which integrates the structured problem-solving approach of the Planning Domain Definition Language (PDDL) \cite{aeronautiques1998pddl} with the strong reasoning ability of large language models (LLMs) to facilitate long-horizon task allocation and execution in heterogeneous multi-robot systems. 
The LLM component of LaMMA-P identifies and allocates sub-tasks based on each robot’s skills, generates PDDL problem descriptions for each robot’s domain, and improves its ability to generalize over human instructions, even when they are vague. These problems are processed by the Fast Downward planner \cite{helmert2006fast} to generate plans for each robot’s assigned sub-task. 
If the initial plan fails, LLMs regenerate and adapt the plan until a viable solution is produced. 
The sub-plans are then combined and converted into executable commands for the robots, allowing for seamless task execution. 
By merging the PDDL’s heuristic search approach with flexible LLM reasoning, LaMMA-P enhances task execution in long-horizon tasks, enabling effective planning in multi-robot systems.

The main contributions of this paper are as follows:
\begin{itemize}
    \item We introduce \textbf{LaMMA-P}, a novel framework that integrates the reasoning ability of large language models (LLMs) with the heuristic planning algorithms of PDDL planners to address long-horizon task planning for heterogeneous robot teams. 
    To the best of our knowledge, LaMMA-P is the first approach that integrates PDDL with LLMs to address multi-agent task planning problems with a flexible number of agents.
    \item We develop a modular design that allows seamless integration of LLMs, PDDL planning systems, and simulation environments, which enables flexible task decomposition and the efficient allocation of sub-tasks based on the skills and capabilities of each robot.
    \item We create \textbf{MAT-THOR}, a benchmark of multi-agent complex long-horizon tasks based on the AI2-THOR simulator, which evaluates the effectiveness and robustness of multi-agent planning methods by providing a standardized set of tasks and performance metrics for long-horizon task execution. Our method achieves state-of-the-art (SOTA) performance on this benchmark in terms of success rate and efficiency.
\end{itemize}

%% file: contents/02relatedwork.tex
Long-horizon planning involves solving complex tasks that require a series of decisions over an extended period. 
Traditional methods such as hierarchical task networks \cite{georgievski2014overview}, general-purpose planning systems based on PDDL \cite{jiang2019task}, and Monte Carlo tree search \cite{browne2012survey} have been employed to address long-horizon problems by decomposing tasks into subtasks or sampling possible future outcomes. 
However, these methods often struggle to scale to larger problem spaces, and they encounter difficulties in generalization and efficiency.
Reinforcement learning \cite{kraemer2016multi} also faces similar challenges when learning transferable and generalizable policies.

Recent studies integrate LLMs into long-horizon planning to tackle these challenges \cite{zhang2023building, nayak2024long, wang2024safe}. 
LLMs can process natural language tasks and translate them into structured formats, which enables more adaptable and context-aware planning. 
For example, recent LLM-enhanced planning approaches rely on translating natural language tasks into Planning Domain Definition Language (PDDL) patterns \cite{liu2023llm+,silver2024generalized,mahdavi2024leveraging,zhou2024isr,dagan2023dynamic,guan2023leveraging,valmeekam2024planbench,xie2023translating}, a framework used to represent and solve complex planning problems. 
Silver et al. \cite{silver2024generalized} and Zhou et al. \cite{zhou2024isr} combine classic PDDL validators with LLM-based chain-of-thought planning to create an automatic loop that enables iterative correction of planning mistakes by LLMs. Singh et al. \cite{singh2024twostep} propose a teacher-student pipeline where two agents guide the planner. 
However, existing approaches are only restricted to single-agent or two-agent systems. 
In contrast, our work advances the field by applying PDDL to LM-driven task planning with an arbitrary number of agents.

%% file: contents/03problemformulations.tex
We consider a scenario where multiple robots collaboratively complete everyday activities in a household environment, such as preparing meals or rearranging objects. 
A human provides high-level natural language instructions, which may lack detailed specifications of required actions. 
This requires task parsing and reasoning, long-horizon planning, and task allocation among robots, which involves identifying necessary sub-tasks from the instructions and determining whether certain sub-tasks can be executed in parallel.

We formalize this problem as a cooperative Multi-Agent Planning (MAP) task \cite{torreno2017cooperative} where multiple agents collaboratively generate a joint task plan to achieve a shared goal. 
Formally, the MAP task is represented as a tuple $\langle \mathcal{AG}, \mathcal{D}, \{\mathcal{A}^i\}^n_{i=1}, \mathcal{P}, \mathcal{I}, \mathcal{G} \rangle$, where $\mathcal{AG}$ is a set of $n$ agents. Each agent $i$ operates within a domain $d_i \in \mathcal{D}$ with its own set of actions $\mathcal{A}^i$. $\mathcal{P}$ is the set of atoms representing the world state, $\mathcal{I} \subseteq \mathcal{P}$ is the initial state, and $\mathcal{G} \subseteq \mathcal{P}$ defines the goal state.
A solution plan is an ordered sequence of actions $\Pi_g = \{\Delta, \prec \}$, where $\Delta \subseteq \mathcal{A}$ are actions and $\prec$ defines their order. The generated plan starts from the initial state $\mathcal{I}$ and leads to the goal state $\mathcal{G}$.

%% file: contents/04methods.tex
Our framework, LaMMA-P, is designed to address long-horizon tasks for heterogeneous multi-agent systems. Fig.~\ref{fig:framework} provides a detailed overview of the modular structure of our framework, which employs language models (LMs) and PDDL planners within six key modules: \textit{Precondition Identifier} (\textit{P}), \textit{Task Allocator}, \textit{Problem Generator} (\textit{G}), \textit{PDDL Validator}  (\textit{V}), \textit{Fast Downward/LLM Planner}, and \textit{Sub-Plan Combiner}. For detailed prompts for each module, refer to our project website or the supplemental video.
Each module serves a specific role in the planning and execution of long-horizon tasks across multiple robots with different skill sets.

We focus on deterministic, fully observable planning tasks.
A PDDL \textit{domain} consists of a name, types, predicates, and operators. Each robot type has a pre-defined domain for its available actions. The domain defines two types: \texttt{robot} and \texttt{object}. One predicate is \texttt{(?o - object)}, where \texttt{?o} is a placeholder for an object. Operators in the domain represent specific robot skills, and all the elements (types, predicates, operators, and objects) have human-readable names. For example, the ``pick up" operator is represented as:
\vspace{-1mm}
\begin{tcolorbox}[colback=gray!20!white, colframe=gray!10!white, width=0.485\textwidth, arc=3mm, auto outer arc, sharp corners]
{\renewcommand{\baselinestretch}{0.8} \footnotesize
\vspace{-2mm}
\begin{verbatim}
(:action PickupObject
  :parameters (?robot - robot 
               ?object - object 
               ?location - object)
  :precondition (and 
                  (at-location ?object 
                   ?location)
                  (at ?robot ?location)
                  (not(inaction ?robot)))
  :effect (and
            (holding ?robot ?object)
            (not(inaction ?robot)))
)
\end{verbatim}
\vspace{-4mm}}
\end{tcolorbox}
\vspace{-4mm}
\subsection{Task Decomposition and Precondition Identifier}
The first step in our framework involves decomposing the task into sub-tasks\cite{kannan2023smart} and introducing a \textit{Precondition Identifier} for each one. Classical planner computes a heuristic \( h(\mathcal{I}, \mathcal{G}) \) by ignoring delete effects \cite{helmert2006fast}, whereas LLMs tackle tasks through probabilistic reasoning over action sequences rather than explicit heuristics.
The relaxed plan heuristic is
\[
h(\mathcal{I}, \mathcal{G}) = \min_{\Pi \in \Pi(\mathcal{I}, \mathcal{G})} \left( \sum_{a \in \Pi} \text{cost}(a) \right),
\]
where only add effects of the actions taken are considered \cite{helmert2006fast}. 
The heuristic \( h(\mathcal{I}, \mathcal{G}) \) estimates the cost of reaching the goal state \( \mathcal{G} \) from the initial state \( \mathcal{I} \), and the function $\text{cost}(a)$ is the cost of performing action $a$. $\Pi$ represents a valid action sequence, and the minimization is taken over all valid action sequences $\Pi(\mathcal{I}, \mathcal{G})$. For LLMs, we define a probabilistic distribution over action sequences:
\[
p\left(a_1, \dots, a_n \mid \mathcal{I}, \mathcal{G}\right) = \prod_{i=1}^{n} p(a_i \mid a_1, \dots, a_{i-1}, \mathcal{I}, \mathcal{G}),
\]
where \( a_1, \dots, a_n \) is the sequence of actions to achieve \( \mathcal{G} \).
The \textit{Precondition Identifier}, similar to relaxed plan, simplifies the preconditions \( P_a \) and effects \( E_a \) by negating unessential effects. This reduction helps LLMs focus on generating action sequences with fewer constraints. The modified probability distribution becomes $p(a \mid P'_a, E'_a, \mathcal{I}, \mathcal{G})$, where the action \( a \) is conditioned on simplified preconditions.   \( P_a \) is the preconditions, and \( E_a \) refers to the effects of each action \( a \). \( P'_a \subseteq P_a \) and \( E'_a \subseteq E_a \) are the reduced subsets, where \(P'_a \) and \( E'_a  \) are outcomes of the optimal probabilistic distribution, eliminating the need to classify the involved objects types.

The LLMs' output can be viewed as the result of applying heuristics as shown in Fig.~\ref{fig:framework}. It approximates the most probable sequence of actions to achieve a given goal $\mathcal{G}$ from the initial state $\mathcal{I}$.
The expected action sequence, given the probabilistic distribution over actions, is defined as
\[
\hat{h}(\mathcal{I}, \mathcal{G}) = \mathbb{E} \left[ \sum_{a} \text{cost}(a) \mid p(a_1, \dots, a_n \mid  P'_a, E'_a, \mathcal{I}, \mathcal{G}) \right],
\]
where \( \hat{h}(\mathcal{I}, \mathcal{G}) \) represents the result of applying heuristics, and \( p(a_1, \dots, a_n \mid  P'_a, E'_a,  \mathcal{I}, \mathcal{G}) \) refers to the probability distribution over all possible action sequences $a_1, \dots, a_n$ conditioned on simplified preconditions \( P'_a \) and effects \( E'_a \).  

In LaMMA-P, LLMs output the Precondition List, which often identifies sub-goals or action sequences more optimally than relaxed planning. The generated initial state is often flawed and thereby significantly impacts the performance of classic PDDL translators \cite{liu2023llm+}. By simplifying preconditions, LLMs can more effectively generate an optimal action path, minimizing unnecessary steps and focusing on the core elements needed to reach the goal. The identified preconditions \( P'_a \) generated by \textit{P} can serve as a grounded sequence of action for generation and validation.

Implementing the few-shot filling prompt below, the LLM generates a sequence of actions along with their associated \( P_a \) and \( E_a \). These preconditions can serve as sub-goals that guide the planning process. The sequence generated by the LLM is then grounded and validated by other modules within the system. For instance, this list of actions and preconditions assists in the generation of the PDDL problem file, providing a structured representation of the task that is compatible with the traditional PDDL-based Fast Downward planner \cite{helmert2006fast}. 

\begin{tcolorbox}[colback=blue!10!white, colframe=blue!10!white, width=0.485\textwidth, arc=3mm, auto outer arc]
\fontsize{9}{11}\selectfont

\vspace{-0.2cm}
\textbf{System Prompt:} You are a task precondition identifier planning for $\mathcal{AG}$ robots. Identify the sub-tasks and action preconditions that are similar to the examples below. 

\textbf{User Prompt:} Example 1: Independent subtasks:

Subtask 1: Put egg in the fridge. (skills required: 

GotoObject, PickupObject, PutObject, OpenObject)

Subtask 2: Prepare apple slices. (skills required: ...)

...

Subtask 1: Put the egg in the Fridge

GoToObject: Robot goes to the egg.

- Parameters: ?robot, ?egg

- Preconditions: (not (inaction ?robot))

- Effects: (at ?robot ?egg), (not (inaction ?robot))

PickupObject: Robot picks up the egg.

...

Subtask 2: Prepapre apple slices

GotoObject: Robot goes to the apple.

...

Example 2: ...
\vspace{-0.2cm}
\end{tcolorbox}

\vspace{-0.2cm}
\subsection{Task Allocation}
Once the task is decomposed into smaller, manageable sub-tasks for a single robot, the sub-tasks are then allocated to heterogeneous multi-agent systems.
The \textit{Task Allocator} module parses the task description, identifying the necessary actions and matching them with the appropriate robots based on their skills and capacities. This allocation process ensures that resources are used efficiently, with the option to execute tasks in parallel when feasible to reduce the time for task completion. Due to the length and inherent randomness of the previous result, LLMs fail to generate in the correct format \cite{levy2024same, wei2022chain}.  
Therefore, we design this module to conclude the identified preconditions and the task allocation into a structured summary. The Task Decomposition Analysis, as shown in Fig.~\ref{fig:framework}, is generated by LLMs within this module.

\subsection{PDDL Problem Generation and Validation}
With the sub-tasks allocated, the next phase involves generating PDDL problems. Each sub-task is translated into a PDDL problem, considering the current state of the environment and the specific goals that need to be achieved.

The \textit{Problem Generator} module constructs the PDDL problems by specifying the relevant objects, actions, initial conditions, and goals. These problems are then passed to the \textit{PDDL Validator} module, which verifies the correctness of the problem files by checking their format and structure to ensure they are ready for the planning phase. 
A PDDL \textit{problem} is characterized by a domain, a set of objects, an initial state, and a goal. An object is identified by a name and a type, e.g., \texttt{Egg - object}. A ground atom is a predicate and a tuple of objects of the appropriate types, e.g., \texttt{(cooked Egg)}. A state consists of a conjunction of true ground atoms, assuming all other ground atoms are false. A goal is a conjunction of ground atoms that must be true in any \textit{goal state}.  For example, in ``Prepare plate with egg", the goal is \texttt{(at-location Egg Plate)}. The full PDDL Problem 1, as illustrated in Fig.~\ref{fig:framework}, is written as
\begin{tcolorbox}[colback=gray!20!white, colframe=gray!10!white, width=0.485\textwidth, arc=3mm, auto outer arc, sharp corners]
{\renewcommand{\baselinestretch}{0.8} \footnotesize
\vspace{-2mm}
\begin{verbatim}
(define (problem prepare-plate-with-egg)
  (:domain robot2)
  (:objects
    Robot2 - robot
    Egg Plate Location1 Location2 - object
  )
  (:init
    (at Robot2 InitLoaction)
    (at-location Egg Location1)
    (at-location Plate Location2)
    (not (inaction Robot2))
  )
  (:goal
    (and(at-location Egg Plate)
      (not (holding Robot2 Egg))
      (not (holding Robot2 Plate)))
  )
)
\end{verbatim}
\vspace{-4mm}}
\end{tcolorbox}
\subsection{Planning and Validation}
Once the PDDL problems are generated, they are passed to the \textit{Fast Downward/LLM Planner} module for plan generation. The Fast Downward planner \cite{helmert2006fast}, operating within the types, constraints, and operators defined in the PDDL domain, produces a sequence of actions that achieve the specified sub-task goals. 
A validator verifies the plan against the robot domain's constraints by analyzing the logs generated by the Fast Downward planner. If the plan is invalid, the system falls back to the LLMs for potential re-planning or refinement, ensuring that a viable solution is available.

\subsection{Sub-Plan Combination and Task Execution} 
After individual sub-plans are generated and validated, the \textit{Sub-Plan Combiner} leverages LLMs with tailored prompts to integrate them into a cohesive final plan that addresses the entire high-level long-horizon task, as shown in Fig.~\ref{fig:framework}. 
The combiner accounts for task parallelizability, scheduling simultaneous execution where possible, while ensuring sequential execution for tasks with dependencies. 
This synchronization ensures smooth transitions between sub-tasks and maintains inter-task dependencies, particularly in multi-robot scenarios. 
By balancing parallel and sequential execution, the system maximizes efficiency throughout task execution.
Once the final plan is combined, the \textit{Plan-to-Code Converter} transforms it into executable code via regular expression matching. The generated code is then executed in the \textit{Scenario Simulator}, which visualizes the robots performing their assigned tasks, completing the task execution process.

%% file: contents/05experiments.tex
\begin{figure*}[!tbp]
	\centering
    \includegraphics[width=0.95\linewidth]{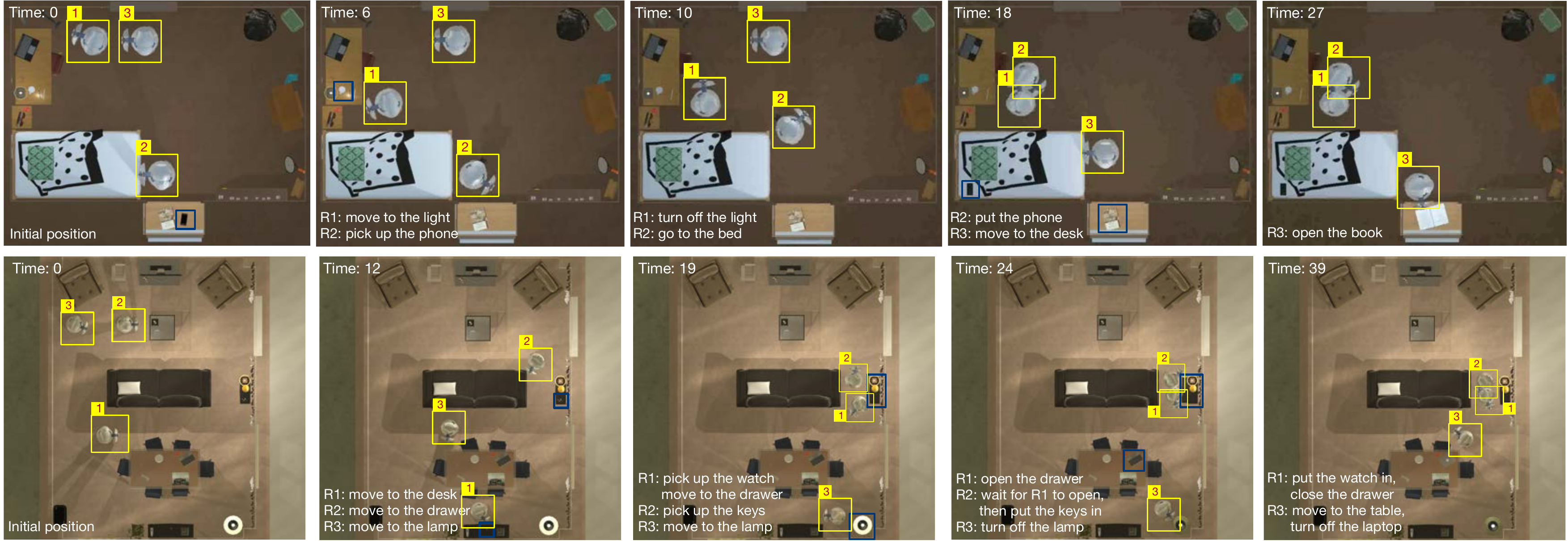}
    \vspace{-0.3cm}
	\caption{The keyframes depict two tasks in different AI2-THOR rooms, highlighting key execution phases. Each row represents a distinct task, with three robots in yellow boxes labeled by number. Objects to be manipulated appear in blue boxes in the prior frame. The first task involves turning off the lights, placing a phone on the bed, and leaving a book open. The second requires placing keys and a watch in a drawer and turning off the lamp and laptop.}
    \vspace{-0.5cm}
	\label{fig:visualization}
\end{figure*}
\subsection{Benchmark Dataset}
We present MAT-THOR, a multi-agent long-horizon task dataset expanded from the SMART-LLM benchmark \cite{kannan2023smart}, to evaluate LaMMA-P and baseline methods based on the AI2-THOR simulator \cite{kolve2017ai2}, as existing datasets do not pose sufficiently challenging tasks for multi-agent task planning..
The MAT-THOR dataset includes 70 tasks across five floor plans, with increasing complexity and vague commands for a more challenging evaluation. It supports testing on task allocation and execution efficiency with two to four robots of varying skills and provides detailed specifications, including initial states, robot skills, and final conditions for success.

The tasks are categorized into three levels of complexity:
\begin{enumerate}
    \item Compound Tasks: These tasks allow for flexible execution strategies (e.g., sequential, parallel, or hybrid). Each robot possesses the necessary specialized skills to handle its assigned sub-tasks independently, with the number of sub-tasks ranging from two to four.
    \item Complex Tasks: These tasks are specifically designed for heterogeneous robot teams, where individual robots may not process all skills to complete their sub-tasks independently. The number of sub-tasks is six or more.
    \item Vague Command Tasks: These tasks present additional challenges with ambiguous natural language instructions, which require the robots to infer missing details.
\end{enumerate}
Our MAT-THOR dataset includes 30 compound tasks, 20 complex tasks, and 20 vague command tasks to evaluate task decomposition, allocation, and execution efficiency. 

\subsection{Evaluation Metrics and Baselines}

We adopt five evaluation metrics \cite{kannan2023smart}: \textit{Success Rate (SR)}, \textit{Goal Condition Recall (GCR)}, \textit{Robot Utilization (RU)}, \textit{Executability (Exe)}, and \textit{Efficiency (Eff)}. 
For a certain task, successful execution occurs when all task-specific goals are achieved. \textit{SR} represents the ratio between successful executions and the total number of tasks. 
\textit{GCR} is the set difference between the ground truth final state and the achieved final state, normalized by the number of task-specific goals. \textit{RU} calculates the ratio between the total transition count of all successful executions and the total ground truth, measuring how effectively the action sequence is planned. 
\textit{Exe} measures the fraction of actions that can be executed, regardless of task relevance. 
\textit{Eff} captures system efficiency as the ratio between the total action time steps and the ground truth time steps. 
Both \textit{RU} and \textit{Eff} are assessed only on successful executions.

We evaluate LaMMA-P across diverse tasks with different language models, including GPT-4o \cite{achiam2023gpt}, Llama-3.1-8B, and Llama-2-13B \cite{dubey2024llama}. 
We adopt SMART-LLM \cite{kannan2023smart} as the state-of-the-art baseline, with GPT-4 replaced by GPT-4o for a fair comparison. 
Also, we introduce a second baseline using only Chain-of-Thought (CoT) prompting with GPT-4o. For the CoT baseline, we manually translate the generated plans into code and evaluate performance metrics accordingly.

\begin{table*}[!tbp]
\centering
\footnotesize
\caption{Evaluation of LaMMA-P and baselines on different categories of tasks in the MAT-THOR dataset}
\vspace{-0.28cm}
\begin{tabularx}{\textwidth}{lXXXXlXXXXXlXXXXX}
\toprule
\multirow{2}{*}{\textbf{Methods}} & \multicolumn{5}{c}{\textbf{Compound}} & \multicolumn{5}{c}{\textbf{Complex}} & \multicolumn{5}{c}{\textbf{Vague}} \\
\cmidrule(lr){2-6} \cmidrule(lr){7-11} \cmidrule(lr){12-16}
 & \textbf{SR~$\uparrow$} & \textbf{Exe~$\uparrow$} & \textbf{GCR~$\uparrow$} & \textbf{RU~$\uparrow$} & \textbf{Eff~$\uparrow$} 
 & \textbf{SR~$\uparrow$} & \textbf{Exe~$\uparrow$} & \textbf{GCR~$\uparrow$} & \textbf{RU~$\uparrow$} & \textbf{Eff~$\uparrow$} 
 & \textbf{SR~$\uparrow$} & \textbf{Exe~$\uparrow$} & \textbf{GCR~$\uparrow$} & \textbf{RU~$\uparrow$} & \textbf{Eff~$\uparrow$} \\
\midrule
CoT (GPT-4o) & 0.32 & 0.67 & 0.40 & 0.72 & 0.59 & 0.00 & 0.55 & 0.12 & 0.47 & 0.38  & 0.00 & 0.24 & 0.00 & 0.00 & 0.00\\
SMART-LLM (GPT-4o) \cite{kannan2023smart} & 0.70 & 0.96 & 0.82 & 0.78 & 0.64 & 0.20 & 0.72 & 0.33 & 0.65 & 0.56 & 0.00 & 0.68 & 0.06 & 0.44 & 0.42  \\
Ours (Llama 2-13B) & 0.36 & 0.88 & 0.45 & 0.84 & 0.54 & 0.05 & 0.78 & 0.08 & 0.51 & 0.43 & 0.00 & 0.66 & 0.00 & 0.00 & 0.00  \\
Ours (Llama 3.1-8B) & 0.45 & 0.92 & 0.53 & 0.79 & 0.61 & 0.15 & 0.83 & 0.28 & 0.62 & 0.48 & 0.00 & 0.72 & 0.00 & 0.00 & 0.00  \\
Ours (GPT-4o) & \textbf{0.93} & \textbf{1.00} & \textbf{0.94} & \textbf{0.91} & \textbf{0.90} & \textbf{0.77} & \textbf{1.00} & \textbf{0.83} & \textbf{0.87} & \textbf{0.67} & \textbf{0.45} & \textbf{0.93} & \textbf{0.48} & \textbf{0.71} & \textbf{0.65} \\
\bottomrule
\end{tabularx}
\vspace{-0.2cm}
\label{table:our-eval}
\end{table*}

\begin{table*}[!tbp]
\centering
\caption{Ablation study on different variations of LaMMA-P}
\vspace{-0.28cm}
\begin{tabularx}{\textwidth}{lXXXXlXXXXXlXXXXX}
\toprule
\multirow{2}{*}{\textbf{Methods}} & \multicolumn{5}{c}{\textbf{Compound}} & \multicolumn{5}{c}{\textbf{Complex}} & \multicolumn{5}{c}{\textbf{Vague}} \\
\cmidrule(lr){2-6} \cmidrule(lr){7-11} \cmidrule(lr){12-16}
 & \textbf{SR~$\uparrow$} & \textbf{Exe~$\uparrow$} & \textbf{GCR~$\uparrow$} & \textbf{RU~$\uparrow$} & \textbf{Eff~$\uparrow$} 
 & \textbf{SR~$\uparrow$} & \textbf{Exe~$\uparrow$} & \textbf{GCR~$\uparrow$} & \textbf{RU~$\uparrow$} & \textbf{Eff~$\uparrow$} 
 & \textbf{SR~$\uparrow$} & \textbf{Exe~$\uparrow$} & \textbf{GCR~$\uparrow$} & \textbf{RU~$\uparrow$} & \textbf{Eff~$\uparrow$} \\
\midrule
Ours (w/o \textit{P} \& \textit{V} \& \textit{G} \& $\mathcal{D}$) & 0.50 & 0.93 & 0.50 & 0.77 & 0.79 & 0.10 & 0.93 & 0.10 & \textbf{0.91} & \textbf{0.72} & 0.10 & 0.85 & 0.25 & \textbf{0.91} & \textbf{1.00}\\
Ours (w/o \textit{P} \& \textit{V} \& \textit{G}) & 0.71 & 0.87 & 0.73 & 0.78 & 0.85 & 0.52 & 0.92 & 0.70 & 0.72 &  0.62 & 0.20 & \textbf{0.93} & 0.32 & 0.61 & 0.85\\
Ours (w/o \textit{P} \& \textit{V}) & 0.67 & 0.84 & 0.72 & 0.76 & 0.89 & 0.58 & 0.87 & 0.60 & 0.77 & 0.61 & 0.15 & 0.87 & 0.22 & 0.86 & 0.77\\
Ours (w/o \textit{P}) & 0.79 & 0.91 & 0.85 & 0.87 & 0.86 & 0.68 & 0.82 & 0.74 & 0.76 & 0.63 & 0.25 & 0.91 & 0.32 & 0.75 & 0.74\\
Ours & \textbf{0.93} & \textbf{1.00} & \textbf{0.94} & \textbf{0.91} & \textbf{0.90} & \textbf{0.77} & \textbf{1.00} & \textbf{0.83} & 0.87 & 0.67 & \textbf{0.45} & \textbf{0.93} & \textbf{0.48} & 0.71 & 0.65 \\
\bottomrule
\end{tabularx}
\vspace{-0.6cm}
\label{table:ablation-variations}
\end{table*}
\subsection{Results and Discussion}
We evaluate LaMMA-P and baseline methods on the MAT-THOR dataset across three distinct task categories: Compound, Complex, and Vague Command. LaMMA-P consistently achieves superior performance compared to the baseline methods across all task categories.

\textbf{Qualitative Analysis.} 
Fig.~\ref{fig:visualization} visualizes task executions in the AI2-THOR simulator (task descriptions are detailed in the caption). Each row presents key execution frames for a task. 
In the first Compound task, Robots 1 and 2 work simultaneously, while Robot 3 starts only after Robot 2 leaves the desk, optimizing parallelism. In the second Complex task, only Robot 1 possesses the ability to open and close objects, creating a dependency between the sub-tasks. Robot 2 waits by the drawer for Robot 1 to arrive and open it. These demonstrate our method's ability to effectively manage complex dependencies while ensuring smooth coordination among the robots for efficient task completion.

\textbf{Quantitative Analysis.} Ours (GPT-4o) improves the average \textit{SR} and \textit{Eff} by 105\% and 36\%, respectively, compared to the strongest baseline SMART-LLM (GPT-4o), which indicates a substantial improvement in task completion, efficiency and generalizability.
These improvements result from LaMMA-P's integration of LMs' advanced reasoning abilities with traditional heuristic search. By accurately decomposing tasks, efficiently assigning sub-tasks, and leveraging heuristic search for planning, our method consistently outperforms the baselines across all metrics. The quantitative results of our experiments are summarized in Table \ref{table:our-eval}. 

In Compound tasks, Ours (GPT-4o) achieves higher \textit{SR} and \textit{Eff} than SMART-LLM (GPT-4o). This improvement stems from Ours (GPT-4o)'s effective subtask identification and allocation, which also leads to higher \textit{Exe} and \textit{GCR}. 
Similarly, Ours (GPT-4o) achieves a significantly higher \textit{SR} than SMART-LLM (GPT-4o), with \textit{Eff} rising more in Complex tasks. This highlights LaMMA-P's strength in handling long-horizon planning, with its advantage becoming more pronounced as task complexity increases, consistently surpassing other approaches under challenging conditions.  The superior performance is further supported by improvements in \textit{Exe}, \textit{GCR}, and \textit{RU}, implying enhanced multi-robot coordination.
For Vague Command tasks, although Ours (GPT-4o)'s \textit{SR} drops, the baseline methods fail on all test cases, demonstrating LaMMA-P's stronger reasoning ability to transform natural language into sub-tasks even with vague instructions. 
Moreover, its generalizability across diverse task descriptions strengthens its advantage in complex environments.
Benefiting from the structured domains of robot skills, \textit{Exe} remains high, indicating that LaMMA-P consistently assigns tasks to capable robots. Ours (Llama 2-13B) and Ours (Llama 3.1-8B) consistently outperform CoT (GPT-4o), demonstrating the effectiveness of our approach in handling complex reasoning tasks, even when compared to larger, state-of-the-art models. 
Both underperform compared to SMART-LLM (GPT-4o) on Compound tasks, likely due to the smaller model size of LLMs. However, as task complexity increases to Complex, Ours (Llama 3.1-8B)'s performance closely approaches that of SMART-LLM (GPT-4o), demonstrating the robustness of our method when tackling more challenging tasks despite the smaller model size.

\textbf{Ablation Study.} We conduct an ablation study to evaluate the impact of various components of LaMMA-P on its overall performance. The results, shown in Table \ref{table:ablation-variations}, indicate that the addition of key elements, such as the pre-defined domains ($\mathcal{D}$), \textit{Problem Generator (G)},  \textit{Precondition Identifier (P)}, and \textit{PDDL Validator (V)}, leads to a significant increase in task performance on both compound and complex tasks. 
The first variation excludes all major components: \textit{P}, \textit{V}, \textit{G}, and $\mathcal{D}$. As a result, performance is significantly lower. The absence of pre-defined robot domains significantly decreases the LLM's ability to identify appropriate sub-tasks for each robot, resulting in incorrect task allocation and sub-optimal performance. After adding $\mathcal{D}$, the \textit{SR} increases substantially for all categories of tasks. This improvement highlights the importance of domain knowledge in guiding LLMs to assign tasks more appropriately to robots. 
Next, the addition of \textit{G} further improves performance by providing a structured approach to describe sub-tasks, making them easier to assign. \textit{Eff} for compound tasks and \textit{RU} for complex and vague command tasks show an increase, indicating more effective task decomposition and improved robot utilization.
\textit{V} brings notable improvements in task performance. It ensures that all generated plans are executable before task execution, which reduces task failures and optimizes the overall workflow.
When all components are included, Ours (GPT-4o) achieves the best performance. The addition of \textit{P} simplifies complex preconditions, enabling LLMs to generate action sequences with fewer constraints to guide further task planning. 

\vspace{-1mm}

%% file: contents/06conclusions.tex
We introduce LaMMA-P, a Language Model-Driven Multi-Agent PDDL Planner that addresses long-horizon task allocation and planning for heterogeneous multi-robot systems. By unifying the strong reasoning ability of LLMs with heuristic search planning based on PDDL, LaMMA-P significantly improves task success rates and robot utilization efficiency over existing methods, while exhibiting generalization across a wide range of tasks. 
LaMMA-P outperforms the strongest baseline SMART-LLM on the MAT-THOR benchmark with a 105\% higher success rate and 36\% higher efficiency in long-horizon tasks. The integration of LLMs enables flexible task translation and assignments, enhancing the system's generalizability across various tasks. While LaMMA-P shows promising results, it assumes fully observable, static environments, which may not always satisfy real-world conditions. Future work may focus on incorporating vision-language models for improved perception and developing adaptive re-planning for dynamic scenarios.